# Player co-modelling in a strategy board game: discovering how to play fast

Dimitris Kalles

Hellenic Open University, Sachtouri 23, 26222, Patras, Greece

`kalles@eap.gr`

**Abstract.** In this paper we experiment with a 2-player strategy board game where playing models are evolved using reinforcement learning and neural networks. The models are evolved to speed up automatic game development based on human involvement at varying levels of sophistication and density when compared to fully autonomous playing. The experimental results suggest a clear and measurable association between the ability to win games and the ability to do that fast, while at the same time demonstrating that there is a minimum level of human involvement beyond which no learning really occurs.

## Statement

This paper has not been published elsewhere and has not been submitted for publication elsewhere.

The paper shares some setting-the-context paragraphs with two referenced papers (Kalles & Kanellopoulos, 2001; Kalles & Ntoutsi, 2002) and a further submission by the same author (Kalles, 2006) with which it is neither identical nor similar. Said submission is currently under review and will be made available on demand to editors for inspection purposes.

No methodology, experiments or results are duplicated. All game data have been recorded for potential examination and reproducibility tests. These data, along with the game code, are available on demand for academic research purposes.

# 1. Introduction

Several machine learning concepts have been tested in game domains, since strategic games offer ample opportunities to automatically explore, develop and test winning strategies. The most widely publicised results occurred during the 1990s with the development of Deep Thought and Deep Blue by IBM but the seeds were planted as early as 1950 by Shannon (1950) who studied value functions for chess playing by computers. This was followed by Samuel (1959) who created a checkers program and, more recently, by Sutton (1988) who formulated the TD($\lambda$) method for temporal difference reinforcement learning (RL). TD-Gammon (Tesauro, 1992; 1995) was the most successful early application of TD($\lambda$) for the game of backgammon. Using RL techniques and after training with 1.5 million self-playing games, a performance comparable to that demonstrated by backgammon world champions was achieved.

Implementing a computer's strategy is the key point in strategy games. By the term *strategy* we broadly mean the selection of the computer's next move considering its current situation, the opponent's situation, consequences of that move and possible next moves of the opponent. In our research, we use a strategy game to gain insight into how we can evolve game playing capabilities, as opposed to programming such capabilities (using mini-max, for example). Although the operational goal of achieving improvement (measured in a variety of ways) is usually achieved in several experimental settings (Ghory, 2004; Osman & Mańdziuk, 2005), the actual question of which training actions help realize this improvement is central if we attempt to devise an optimized training plan. The term *optimize* reflects the need to expend judiciously the training resources, be it computer power or human guidance.

This paper attempts to define quantitative metrics of the quality of playing models based on knowledge of how these models were evolved. It builds on previous experimental research and assumes that computer players receive very simple feedback during the game; namely, a lump credit (penalty) for eventual win (loss) and a scaled credit (penalty) for the capturing of pawns. The playing models are evolved based on gaming sessions with varying human involvement and varying human skill. The encouraging result is that even with such limited feedback, the successful models seem to have also developed the capability to limit the amount of moves they need in order to secure a win, even though no such quantitative information has been built into the credit scheme. We consider this to be a strong indicator of the potential of our approach, namely that playing speed has emerged as an important quality metric despite its absence from the design stage. We argue that this is important in the design of any knowledge-based system that deals with strategy issues.

The rest of this paper is organised in four sections. The next section presents the basic details of the game and its introductory analysis. The third section describes our experimentation on training. The fourth section discusses the impact and the limitations of our results and examines the potential for developing autonomous players based on very low information content. The concluding section summarises the work.

## 2. Game Description and Analysis – in brief

The game is played on a square board of size $n$, by two players. Two square bases of size $a$ are located on opposite board corners. The lower left base belongs to the white player and the upper right base belongs to the black player. At game kick-off each player possesses $\beta$ pawns. The goal is to move a pawn into the opponent's base.

The base is considered as a single square, therefore every pawn of the base can move at one step to any of the adjacent to the base free squares (see Fig. 1 for examples and counterexamples of moves). A pawn can move to an empty square that is vertically or horizontally adjacent, provided that the maximum distance from its base is not decreased (so, backward moves are not allowed). Note that the distance from the base is measured as the maximum of the horizontal and the vertical distance from the base (and not as a sum of these quantities). A pawn that cannot move is lost (more than one pawn may be lost in one round). If some player runs out of pawns he loses.

The leftmost board in Figure 1 demonstrates a legal and an illegal move (for the pawn pointed to by the arrow). The rightmost boards demonstrate the loss of pawns (with arrows showing pawn casualties), where "trapped" pawn automatically draws away from the game. As a by-product of this rule, when there is no free square next to the base, the rest of the pawns of the base are lost.

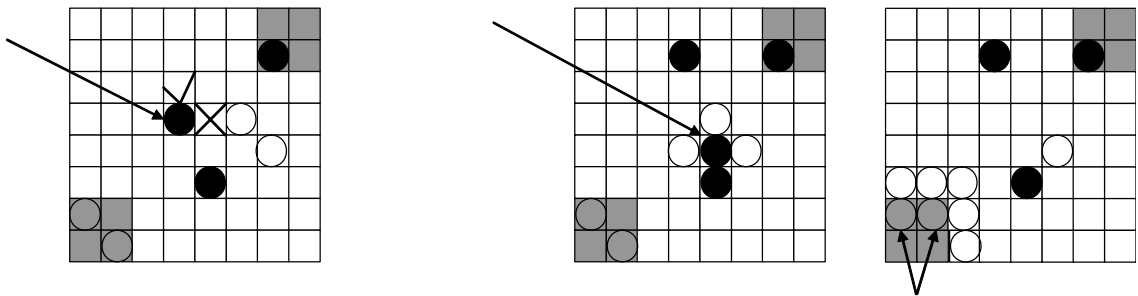

**Figure 1: Examples and counterexamples of moves**

The game is a discrete Markov procedure, since there are finite states and moves. The *a priori* knowledge of the system consists of the rules only. The agent's goal is to learn an optimal policy that will maximize the expected sum of rewards in a specific time, determining which action should be taken next given the current state of the environment. A commonly used starting *ε*-greedy policy with *ε=0.9* was adopted, i.e. the system chooses the best-valued action with a probability of 0.9 and a random action with a probability of 0.1. At the beginning all states have the same value except for the final states. After each move the values are updated through TD(*0.5*).

Each player approximates its state space with a neural network (Sutton & Barto, 1998). The input layer nodes are the board positions for the next possible move (in total $n^2-2a^2+10$). The hidden layer consists of half as many hidden nodes. There is one output node; it stores the probability of winning when one starts from a specific game-board configuration and then makes a specific move.

Note that, drawing on the above and the game description, we conclude that we cannot effectively learn a deterministic optimal policy. Such a policy does exist for the game (Littman, 1994), however the use of an approximation effectively rules out such learning. Of course, even if that was not the case, it does not follow that converging to such a policy is computationally tractable (Condon, 1992).

## 3. Experimentation – Planning and Implementation

Earlier experimentation (Kalles & Kanellopoulos, 2001) initially demonstrated that, when trained with self-playing games, both players had nearly equal opportunities to win and neither player enjoyed a pole position advantage. Follow-up research (Kalles & Ntoutsi, 2002) furnished preliminary results that suggested a computer playing against itself would achieve weaker performance when compared to a computer playing against a human player. More recently we focused on the measurable detection of improvement in automatic game playing, by constraining the moves of the human (training player), while experimenting with different options in the reward policies (Kalles, 2006).

In the present paper we advance one step in opening up the available moves of the human player and compare the learned playing models to the model corpus we have been building up in this line of research. The outcome is the generation of game metrics that will allow us to develop a better insight into which rounds of experiments we should be planning next in our quest to develop a human-level playing machine with minimal human interaction.

Experiments are reported in terms of game batches. Each game batch consists of 5,000 computer-vs.-computer (CC) games, carried out in 5 stages of 1,000 games each. For batches that have involved human-vs.-computer (HC) games, each CC stage is interleaved with a HC stage of 10 games. Thus, HC batches are 5,050 games long. In HC games, a human is always the white player. Two HC batches have been tested with 1 game per stage to further explore the potential of accelerated learning; these $HC_1$ batches are 5,005 games long. Note also that earlier CC experiments (Kalles, 2006) were initially designed in stages of 10,000 games each; however it was found that shorter stages seem to better avoid the dilution of "lessons learned" by a computer when playing against a human.

We now show the alternatives for the human player in Table 1. Briefly describing them, in Policy 1, we always move from the bottom-left base to the north and then move right, attempting to enter the black base from its vertical edge (see Figure 2, left). In Policy 2 (see Figure 2, right), the white player (human) now attempts to cover all ground of the board in one HC stage, but with a different trajectory each time. For example, in the $5^{th}$ game of a HC stage, the white player may start at the same cell as in the $2^{nd}$ or the $4^{th}$ game and will then proceed along the solid arrowed line. Only if the black player manages to effectively block access to its base will the white player be allowed to use a second pawn out of its base. In the $9^{th}$ and $10^{th}$ game of such a stage, the white player will move two pawns out of its base, and then proceed by moving in the central range of the board (shaded area).

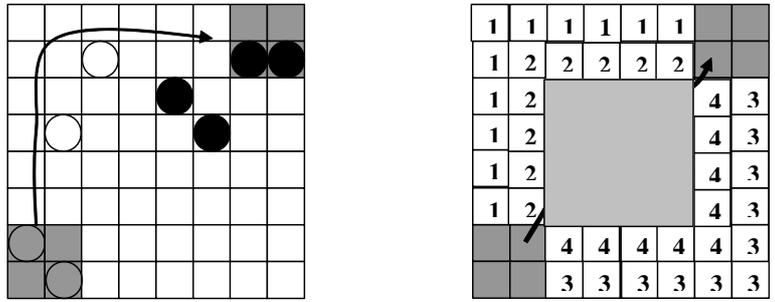

**Figure 2: The path of a human player**

**Table 1. Policies of white human player**

| | White player always starts from bottom-left |
|---|---|
| 1 | North, then Right, attempting to enter from vertical edge |
| 2 | 8 "parallel-like" routes in all board + 2 "central" arbitrary routes |

The reward scheme assigns a credit of 100 for a game win and of -100 for a game loss. It also rewards the capturing of pawns by calculating the difference of pawn counts between the two players and by scaling that difference to a number in [-100,100].

We now show the graph representing the structure of the experimental plan (with reference to Figure 3) and describe the key notations with some examples. All in all we report on 14 game batches, indexed from 9 up to 22.[1]

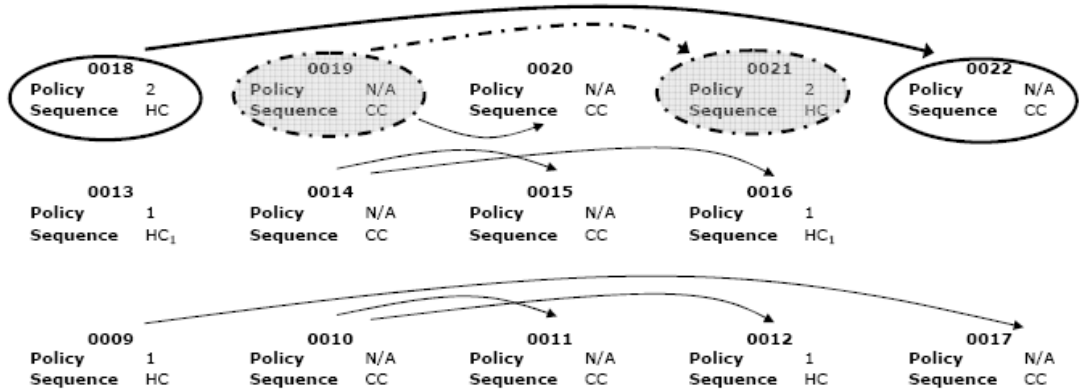

**Figure 3: The experimental plan**

First, consider batch 18 where Policy 2 was used for the human player. After batch 18, the neural networks representing the models for the white and the black player were fed as input to batch 22 and subsequently evolved in CC mode. In contrast, (CC) batch 19 fed its output into (HC) batch 21. So, batches 21 and 22 have been both based on the same number of CC and HC games, albeit in a different order, both having started at a *tabula rasa* point (initialized neural networks).

After concluding the experiments we can measure the relative effectiveness of the learning policies that delivered the model for each one of any batches *X* and *Y*. Each comparison is done in two steps (of 1,000 CC games each) where, in the first step the white player of the $X^{th}$ batch plays again the black player of the $Y^{th}$ batch (in the second step a similar setting applies). A measurement sample is shown in Table 2.

---

[1] For traceability reasons, we start indexing at 9 because we have used the numbers 1..8 for experiments referred to in earlier work.

**Table 2. Comparative evaluation of distinct learning paths between batches *X* and *Y***

|  | Games Won | | Average # of Moves | |
|---|---|---|---|---|
|  | White | Black | White | Black |
| White$_x$ – Black$_y$ | 715 | 285 | 291 | 397 |
| White$_y$ – Black$_x$ | 530 | 470 | 445 | 314 |

### 3.1 Testing the acceleration hypothesis

The first experimental session was designed to test the acceleration hypothesis. We ran a tournament between batches 9, 12, 13 and 16. An extract of the experimental plan showing these batches is shown in Figure 4. Note that batches 9 and 12 are based on the same number of HC experiments, but batch 12 is longer in terms of its CC experiments (since it is already based on a full CC batch). Batches 13 and 16 are similar but accelerated in the sense that they use less HC experiments (1 per stage).

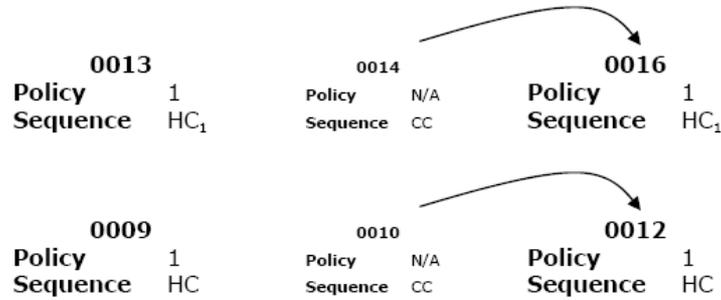

**Figure 4: A short test for the acceleration hypothesis**

In our earlier work (Kalles, 2006) we had presented experimental evidence that scaling down the number of games in a CC stage from 10,000 to 1,000 had produced favourable results in the sense that the extra 9,000 games per stage did not seem to really refine the knowledge of the players. In this work we have now trimmed the HC batches and left the CC ones intact; after all it is human interven-tion that we argue must be judiciously used to improve the computer player. A substantial cost to train a computer player is the time that must be expended by a human trainer.

We have calculated the number of games won by each player in this tournament as well as the average number of moves in the games each player was involved.

Table 3 presents the results for the number of games won. For example, $W_9$ (the white player of batch 9) has won 222 games more than $B_{13}$ during their CC comparison (of 1,000 games). Also, $W_9$ has participated in 3 CC comparisons and, in total, has lost 238 games more than it has won, thus earning it the 3$^{rd}$ place among all white players. In contrast $B_{13}$ has lost 1032 more games than it has won, putting it at the 4$^{th}$ place among all black players (Note that, for a black player, a small negative value -large absolute value- signifies a big win, whereas a positive value signifies a loss).

**Table 3: Testing accelerated learning: number of games won**

|       |      | Black |      |      |      |       |      |
|-------|------|-------|------|------|------|-------|------|
|       |      | 9     | 12   | 13   | 16   | Sum   | Rank |
| White | 9    | N/D   | -148 | **222** | -312 | **-238** | **3** |
|       | 12   | -782  | N/D  | 834  | -234 | -182  | 1    |
|       | 13   | 178   | -206 | N/D  | -166 | -194  | 2    |
|       | 16   | -640  | -540 | -24  | N/D  | -1204 | 4    |
|       | Sum  | -1244 | -894 | **1032** | -712 |       |      |
|       | Rank | 1     | 2    | **4** | 3    |       |      |

Table 4 presents the results for the average number of moves.

**Table 4: Testing accelerated learning: average number of moves**

|  |  | Black | | | | | |
|---|---|---|---|---|---|---|---|
|  |  | 9 | 12 | 13 | 16 | Sum | Rank |
| White | 9 | N/D | 211 | 239 | 314 | 764 | 2 |
|  | 12 | 97 | N/D | 72 | 355 | 524 | 1 |
|  | 13 | 538 | 741 | N/D | 657 | 1936 | 4 |
|  | 16 | 302 | 240 | 380 | N/D | 922 | 3 |
|  | Sum | 937 | 1192 | 691 | 1326 |  |  |
|  | Rank | 2 | 3 | 1 | 4 |  |  |

However, it is when we combine the performance of the white and the black player in each batch that we really obtain an indication of whether one batch has produced good players that have participated in relatively short games. Table 5 presents the results for the average number of moves. While the scale of the experiments is not yet sufficiently large to allow for a steadfast argument (though we cannot quantify what "sufficiently large" may really mean), we do observe a clear association between the ability to play fast and the ability to win games. Nevertheless, this observation is only a side effect of the experimentation, however noticeable. The most significant one is that the two accelerated learning batches based on sparse human involvement are abysmally low in performance. This suggests that there may be a lowest acceptable rate of human involvement, beyond which the effect of human training is effectively diluted.

**Table 5: Testing accelerated learning: aggregate results**

| Batch |
|---|

|  | 9 | 12 | 13 | 16 |
|---|---|---|---|---|
| Total number of games won | 1006 (1) | 712 (2) | -1226 (4) | -492 (3) |
| Average number of moves | 284 (1) | 286 (2) | 438 (4) | 375 (3) |

Going back to the details in Table 3 and in Table 4, we note that white players do in general bad. We have thus confirmed our earlier findings that black players are more effectively trained, since they learn to defend and may then (even, accidentally) win.

It is interesting to see that $W_{13}$ is involved in all three lengthiest matches and that there are two very fast matches, namely $W_{12}$-$B_9$ and $W_{12}$-$B_{13}$ and that $B_9$ and $B_{13}$ have been both generated based on the smallest possible number of CC experiments. This clearly suggests that the unwarranted introduction of CC experiments may result in the development of spurious navigation paths in the state space of the game, leading the player into wandering. It is also very interesting to see that these two very fast matches also result in very clear winning outcomes. That $W_{12}$-$B_{13}$ is the match where a white player scores a significant win is a further confirmation of the fact that the sparseness of human involvement must be judiciously set. Note, for example, that $B_{13}$ has been based on a ratio of 1:1000 for HC and CC games, whereas for $B_9$ the corresponding ratio is 1:100.

**3.2 Identifying efficient and effective learners**

The second experimental session was designed to further test the initial outcome of the previous stage; namely that fast experiments generally indicate the existence of a good player. We used all batches shown in Figure 3 to create tournament couples using batches with human involvement and the results were tabulated using the template shown in Table 2. However, we needed to calculate two new quantities (see Table 6 for an example based on the results for the tournament between batches 9 and 12):

- The *speed ratio* is defined as the ratio of the average number of moves in each round of the tournament. For normalization purposes, the speed ratio must be at least 1.

- The *advantage ratio* is defined as a ratio of two further ratios: each component is the ratio of the number of games won by the white player over the number of games won by the black player. The composite ratio indicates the difference between the two batches. For normalization purposes, the advantage ratio must be at least 1.

**Table 6. Comparative evaluation of distinct learning paths with constructed ratio features**

|  | Average # of Moves | | | |
|---|---|---|---|---|
|  | White | Black | Overall | Speed Ratio |
| $W_9 - B_{12}$ | 258 | 176 | 211 | = 211 / 97 = 2.17 |
| $W_{12} - B_9$ | 337 | 68 | 97 |  |

|  | Games Won | | | |
|---|---|---|---|---|
|  | White | Black | Overall | Advantage Ratio |
| $W_9 - B_{12}$ | 426 | 574 | = 426 / 574 = 0.74 | = 0.74 / 0.12 = 6.07 |
| $W_{12} - B_9$ | 109 | 891 | = 109 / 891 = 0.12 |  |

The two ratios help indicate whether apparently fast games are associated with at least one apparently good batch. We show the results in terms of increasing speed ratio in Figure 5.

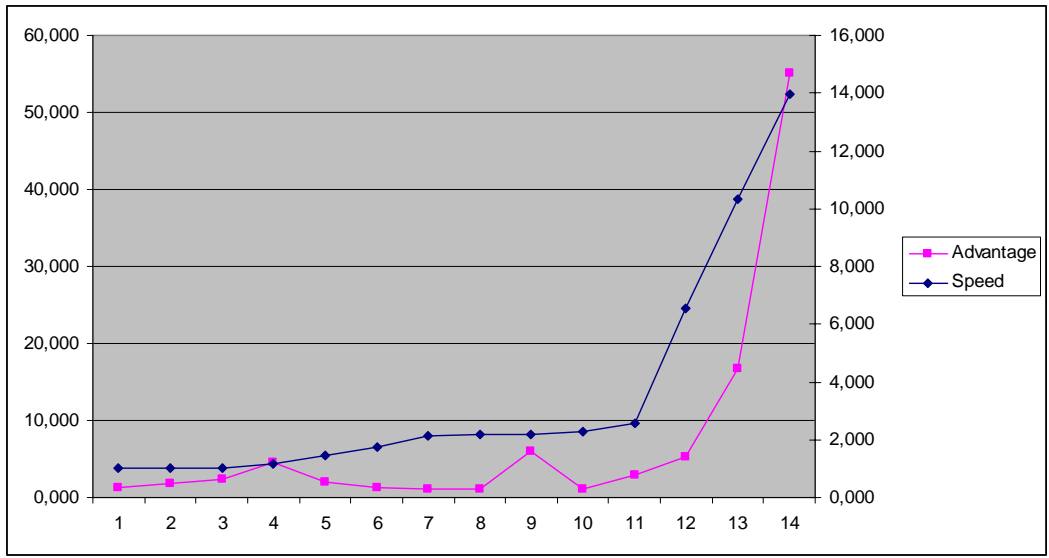

**Figure 5: A comparative presentation of speed and advantage ratios**

A straightforward observation is that, just like in section 3.1, a large speed ratio is associated with a large advantage ratio. There are two spurious measurements in data points 4 and 9, where the advantage ratio is much larger that what the speed ratio would imply. Investigating back into the individual results we noticed that data point 4 refers to the tournament between batches 9 and 12, while data point 9 refers to the tournament between batches 18 and 21. However (see Figure 3 for a reference) these are two similar sets of batches, since they are based on the same ratio of HC vs. CC games, and such games are in the same order. The difference between these data points is that batches 18 and 21 have been both based on Policy 2 ("cover all board") and have had the opportunity to generate many more meaningful paths to be subsequently explored by the CC games. In contract, batches 13 and 16 (their tournament is at data point 6) amply demonstrate that the sparseness of human involvement does not lead to any learned winning behaviour and allows both players to wander in almost the same amount of time (both ratios are at about 1).

For our next attempt at analysis we redefined the advantage ratio towards more "compressed" values. Table 7 shows an example of the new definition. It could be argued that the new composite ratio better reflects the difference between the two batches.

**Table 7. Comparative evaluation of distinct learning paths with a redefined advantage ratio**

| | Games Won | | | Advantage Ratio |
|---|---|---|---|---|
| | White | Black | Overall | |
| $W_9 - B_{12}$ | 426 | 574 | $W_9 B_9 = 426 + 891 = 1317$ | $= 1317 / 683 = 1.93$ |
| $W_{12} - B_9$ | 109 | 891 | $W_{12} B_{12} = 109 + 574 = 683$ | |

Again, we show the results in terms of increasing speed ratio in Figure 6.

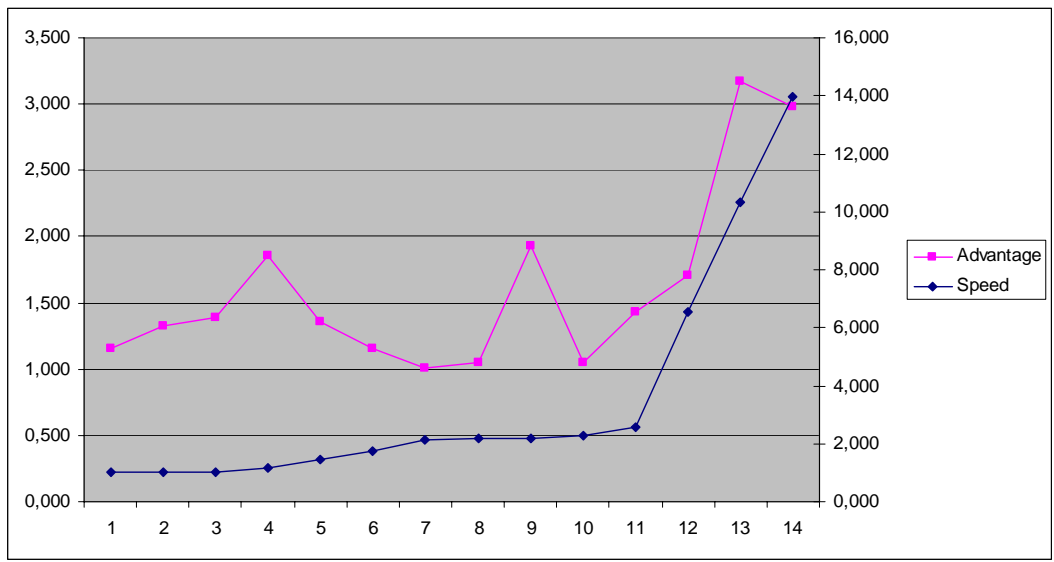

**Figure 6: A comparative presentation of speed and redefined advantage ratios**

It is very interesting to note that data points 4 and 9 demonstrate the same spuriousness under the new definition. We interpret this to be a strong indication of a qualitative similarity in the learned behaviour by these particular playing sequences and configurations.

## 4. Discussion

The cautious reader might ask what may be the ramifications of this experimental research in the design of knowledge-based systems at large, besides being a showcase of application of specific techniques and besides the goal of developing autonomous computer playing. We believe that the fundamental answer lies in the identification of one archetypal design peculiarity of such systems: the elicitation of knowledge.

When confronted with strategy issues in an unknown field we, humans, tend to learn by trial and error. By confronting problems in a new domain we slowly develop measures of success and of measuring up the difficulty of the problem at hand. When on one's own, the selection of problems is a delicate exercise in resource management and can easily lead to exasperation. Much more effective is the employment of a tutor, if we can afford one.

Too coarse problems lead us to learn nothing while too fine problems focus us too much and render us unable to generalize if we are novices. Our research attempts to navigate the fine line between the sparseness and density of learning examples when the computer serves as the student (Rosin, 1997; van den Herik *et al.*, 2005; Long, 2006). Our goal is to establish some examples of successfully tuning the "syllabus" when the computer serves as the student. Along that direction we expect that the length and the content of a training session will slowly become evident if we spend enough time even with very simple feedback (for example, pursuing the analogy, how one does in tests). This can be best visualized in Figure 5 and in Figure 6, where the similarity of data points 4 and 9 is reflected by the structural similarity of the corresponding batches that have generated the player models.

A subtle point therein is that while the ratios are very similar, the average number of moves in data point 9 is about 3 times the average number of moves in data point 4, which probably accounts for the richness in exploration possibilities incurred by Policy 2. Such richness is necessary to improve the standard of the computer player and achieve playability by human standards (Keh & Wang, 2004).

While we have drawn analogues to human tutoring, we are yet far from having a system at a moderately good playing level (the most we have achieved so far is the development of relatively good defensive tactics). We have elsewhere argued (Kalles, 2006) that there are two key technical directions to follow: first, devising an experimentation engine that will produce and perform experimental plans (see Partalas *et al.* (2006) for a related problem) and, second, devising an experiments' life-cycle and workflow administration tool that will support the human expert in designing the learning experiences of the computer. However, both directions do demand even a loose formulation of some type of effectiveness measures for the training sessions. We believe that this paper has reported significant progress in that direction.

We have not yet formulated a theoretical framework to deal with the temporal modelling of human tutoring. Such a framework would probably also need to accommodate an already implemented extension: that of extra knowledge readily available but not used by the white player. Specifically, in HC games, the computer always maintains a model of the white player (played by a human) and such a model obviously always has a recommendation handy for the next white move. Of course, at this early stage of game development, the human player does know better but that knowledge is only reflected in the white player's model at the temporal points where some credit or penalty is involved. While this ensures a relative balance, it slows the game down since the white player model could have been also updated at points where the human player selects a move that is different from the one the model would recommend. However, the improved model of the white player during the HC stage could in principle (as has been seen in the experiments reported in this paper) lead to increased pressure in the black model during the CC stages, leading to more effective learning.

## 5. Conclusion

We have presented experimental evidence that autonomous playing in a strategy board game is a domain where judicious human involvement can lead to more than measurable improvement. The timing and scope (extent) of that involvement can be seen to produce a relative richness in the ability of autonomous players to navigate faster in the state space of available game tactics. That such behaviour emerges is a strongly encouraging result, particularly so since it is not explicitly implemented in the credit scheme of the autonomous players. The price for the emergence of such behaviour is the (up to now) realization that human involvement cannot be simply dispersed throughout an experimental plan but must be carefully and exercised. The analogue to human tutoring is very interesting.

In our earlier work (Kalles, 2006) we had suggested that significant applied research is required for the establishment of tools that will streamline the experimentation process and that such workflow-like tools to assist the human game designer (and, researcher) are more important than the autonomous management of this process (Tesauro, 2002). While we still stand by that claim, we also believe that the establishment of objective and credible success measures will be important in both directions. This paper has reinforced the quality of experimental indices that may serve as such measures.

## Acknowledgements

During the past year Christos Kalantzis, a student of the author, designed and carried out long series of experiments. Numerous conversations on the suitability of these experiments and the interpretations of their results have influenced this work.

Player co-modelling in a strategy board gameH.J. van den Herik, H.H.L.M. Donkers, P.H.M. Spronck (2005). "Opponent Modelling and Commercial Games", *Proceedings of IEEE 2005 Symposium on Computational Intelligence and Games*, Essex University, Colchester, UK. pp 15-25.

D. Kalles and P. Kanellopoulos (2001). "On Verifying Game Design and Playing Strategies using Reinforcement Learning", *ACM Symposium on Applied Computing, special track on Artificial Intelligence and Computation Logic*, Las Vegas.

D. Kalles, E. Ntoutsi (2002). "Interactive Verification of Game Design and Playing Strategies", *IEEE International Conference on Tools with Artificial Intelligence*, Washington D.C.

D. Kalles (2006). "On Measuring the Impact of Human Actions in the Machine Learning of a Board Game's Playing Policies", *submitted paper*.

H.C. Keh, and Y. Wang (2004). "Prolonging the Life Cycle of Computer Games: Taking Fighting Games as an Example: How to Make Virtual Masters More Real?", *Proceedings of Computer Graphics and Imaging Conference*.

M.L. Littman (1994). "Markov Games as a Framework for Multi-Agent Reinforcement Learning", *Proceedings of 11$^{th}$ International Conference on Machine Learning,* San Francisco, pp 157-163.

J. Long (2006). "Game Theoretic and Machine Learning Techniques for Balancing Games". *M.Sc. Thesis,* University of Saskatchewan, Canada.

D. Osman, J. Mańdziuk (2005). " TD-GAC: Machine Learning Experiment with Give-Away Checkers," *Issues in Intelligent Systems. Models and Techniques*, M. Dramiński *et al.* (eds.), EXIT, pp. 131-145.

I. Partalas, G. Tsoumakas, I. Katakis and I. Vlahavas (2006). "Ensembe Pruning Using Reinforcement Learning", *Proceedings of the 4th Panhellenic conference on Artificial Intelligence, Heraklion*, Greece, Springer LNCS 3955, pp. 301-310.Page 18 of 19